\documentclass[journal, 9pt]{IEEEtran}

\ifCLASSINFOpdf
\else
\fi
\usepackage{hyperref}

\usepackage{graphicx}
\graphicspath{ {./images/} }
\usepackage[font={footnotesize ,it}]{caption}
\usepackage{tabularx}
\usepackage{amsmath}
\usepackage{float}
\usepackage{siunitx}
\usepackage{multirow}
\usepackage[utf8x]{inputenc} 

\begin{document}
%
\title{A Deep Learning Framework to Reconstruct Face under Mask}

%
%
%

\author{\IEEEauthorblockN
\Large
{Gourango Modak\IEEEauthorrefmark{1},
Shuvra Smaran Das\IEEEauthorrefmark{2},
Md. Ajharul Islam Miraj\IEEEauthorrefmark{3},
Md. Kishor Morol\IEEEauthorrefmark{4}\\
~\IEEEmembership{}}
\IEEEauthorblockA{\IEEEauthorrefmark{1}Department of Computer Science, 
American International University-Bangladesh}\\
{
\IEEEauthorblockA{
\{
\IEEEauthorrefmark{2}shuvradas59,
\IEEEauthorrefmark{3}miraj.cs18}
\}@gmail.com \\
\{
\IEEEauthorrefmark{1}18-37102-1@student. ,
\IEEEauthorrefmark{4}kishor@,
\}aiub.edu
}
}
\setlength{\textfloatsep}{6pt}

%
%


\markboth{2022 7th Conference on Data Science and Machine Learning Applications (CDMA)}
{Shell \MakeLowercase{\textit{et al.}}: Bare Demo of IEEEtran.cls for IEEE Journals}

%


\maketitle


\begin{abstract}

While deep learning-based image reconstruction methods have shown significant success in removing objects from pictures, they have yet to achieve acceptable results for attributing consistency to gender, ethnicity, expression, and other characteristics like the topological structure of the face. The purpose of this work is to extract the mask region from a masked image and rebuild the area that has been detected. This problem is complex because (i) it is difficult to determine the gender of an image hidden behind a mask, which causes the network to become confused and reconstruct the male face as a female or vice versa; (ii) we may receive images from multiple angles, making it extremely difficult to maintain the actual shape, topological structure of the face and a natural image; and (iii) there are problems with various mask forms because, in some cases, the area of the mask cannot be anticipated precisely; certain parts of the mask remain on the face after completion. To solve this complex task, we split the problem into three phases: landmark detection, object detection for the targeted mask area, and inpainting the addressed mask region. To begin, to solve the first problem, we have used gender classification, which detects the actual gender behind a mask, then we detect the landmark of the masked facial image. Second, we identified the non-face item, i.e., the mask, and used the Mask R-CNN network to create the binary mask of the observed mask area. Thirdly, we developed an inpainting network that uses anticipated landmarks to create realistic images. To segment the mask, this article uses a mask R-CNN and offers a binary segmentation map for identifying the mask area. Additionally, we generated the image utilizing landmarks as structural guidance through a GAN-based network. The studies presented in this paper use the FFHQ and CelebA datasets. This study outperformed all prior studies in terms of generating cutting-edge results for real-world pictures gathered from the web.

\end{abstract}


%
\IEEEpeerreviewmaketitle

\vspace{-.8em}
\section{Introduction}
%
%
%
%
\IEEEPARstart{D}{ue} to the recent pandemic, it has become necessary for everyone to wear masks. Almost everywhere on the globe, it has become mandatory to wear a mask while leaving the house. However, a delinquent may use this rule and conduct some crimes while wearing a mask \cite{r1}. It's very difficult for non-technical individuals to guess who is behind the mask. Take off the mask that protects almost half of one's face, and it's more probable that one's identity will be more readily determined. However, image inpainting is a method for recovering missing or damaged portions of images. It may be used for a variety of purposes, including picture restoration and editing. In comparison to natural landscapes such as seas and lawns, altering faces, which is the subject of this piece, is more difficult. Due to the face complex structure, attribute consistency and topological structure must be maintained to get consistent and realistic results. Otherwise, a minor breach of the reconstructed face's topological structure and/or attribute consistency will almost certainly result in a major perceptual defect. However, although considerable research has been done on image inpainting, the first non-learning methods \cite{r2} and \cite{r3} were developed to remove undesirable items and restore lost data by matching comparable patches from the rest of the picture. 
\begin{figure}[t]
\includegraphics[width=0.49\textwidth]{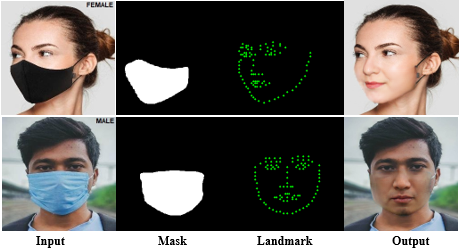}
\centering
\caption{Real-world image results for different face shapes.}
\end{figure}
After identifying comparable patterns in a database consisting of millions of scene photos, they applied them to the damaged area. According to the article, \cite{r4} it is possible to remove eyeglasses from face pictures by using PCA reconstruction and iterative error compensation techniques. These non-learning algorithms, on the other hand, are restricted to the removal of tiny objects from images and are not capable of learning. Image editing algorithms may now enhance their performance by learning from large-scale datasets, outperforming non-learning approaches for the first time when it comes to removing unwanted aspects from pictures, thanks to recent advances in learning-based techniques. This work \cite{r5} demonstrates how to use a GAN setup with two discriminators to remove unwanted objects from a photo and repair damaged regions in a picture. 
Additionally, learning-based techniques for image editing are effective in removing objects with few structural and visual distinctions. However, due to the object's huge size and intricacy, these techniques do not perform well for inpainting the masked faces. For example, masks often conceal not just half of the face's features, but also specific regions outside the face's natural limits. It begins at the tip of the nose (just below the eyes) and extends to certain portions of the neck, as well as some areas beyond the cheeks. A GAN-based network approach guided by facial landmarks is proposed as a solution to this problem, which automatically removes the mask and fills the detected mask region of the face, ensuring that the completed face appears natural and realistic while also being consistent with the other parts of the image. The proposed approach has the following advantages: Face landmarks are used as structural guidance in this research because of their compactness, sufficiency, and long-term durability. Gender classification has proven to be a tough challenge in our past experiments, especially when it comes to image inpainting, and the existing paperwork did not provide the expected result. The consequence of female face inpainting results in a person with more masculine facial features. We have tackled this issue by separating the males and females. Furthermore, we have divided our research problem into three major stages: landmark detection, object identification for the discovered mask area, and image completion for the detected mask region. Just after classifying the gender in the system, we began by detecting the masked face image's landmarks. In the second step, we identified the non-face item, i.e., the mask, and used the Mask R-CNN network to create the binary mask of the observed mask area. To finish the faces in the third stage, we build an inpainting network that follows the anticipated landmarks and makes use of distant spatial context and temporal feature maps to ensure attribute consistency. Additionally, we divided the dataset into two parts, male and female, and trained the inpainting model separately to ensure correct gender attribute consistency. Moreover, since there are no face image pairings with and without mask objects, we generated a matched synthetic dataset by altering photos from the publically accessible FFHQ, CelebA dataset.
\\
The following are the work's main contributions:
\begin{description}
\item[$\bullet$]We've developed an end-to-end method in which we first classify the input image as male or female. Then, we locate the image's landmark and the mask regions where the inpainting model will inpaint the addressed mask region. 
\item[$\bullet$]As facial landmarks represent the topological structure and characteristics of the face such as pose, gender, ethnicity, and expression. We used a robust landmark module to predict landmarks on the masked face image. Also, to maintain the gender attribute correctly, we trained the inpainting module separately on males and females. 
\item[$\bullet$]To complete faces, we create an inpainting module that is guided structurally by the anticipated landmarks. Also, the subnet leverages remote geographical context and links temporal feature maps to ensure attribute consistency.
\item[$\bullet$]Despite being trained on a synthetic dataset, our generator model generates realistic images while maintaining the topological structure and features of the face in challenging real-world images.
\item[$\bullet$]Extensive tests are performed to determine the optimal performance of our design, demonstrate the qualitative comparison to state-of-the-art alternatives, and the quantitative performance of the CelebA, FFHQ dataset.
\end{description}

\vspace{-.8em}
\section{Related Work}
For the image completion part, there were several previous previews. The feasibility of structure propagation depends on the ability to reproduce both texture and structure via exemplar-based texture generation. These articles \cite{r10},\cite{r2} offer a best-first method for propagating confidence in synthetic pixel values like how information propagates during inpainting. Exemplar-based synthesis is used to generate the exact color values. A block-based sampling technique achieves computational efficiency. A new image completion method is suggested in this study \cite{r3}, which is driven by a large collection of photos obtained from the Internet. The algorithm fills in gaps in images by searching the database for comparable image areas that are not smooth. For any input image, the algorithm can produce a different variety of output. This work \cite{r4} detects the regions obstructed by the glasses and achieves a natural-looking smiling picture without eyeglasses. In contrast, learning-based image editing perspectives outperform conventional methods both intuitively and statistically. Object removal is the major application of many prior supervised learning-based image editing studies. In the GFCM \cite{r11} study, they used a neural network to produce information for lacking areas. This technique works best with low-resolution pictures (178 x 218 pixels) and produces artifacts when a damaged area is close to the image's boundaries. To enhance prediction accuracy, the GCA \cite{r6} research takes advantage of surrounding picture characteristics as references throughout the network training process. This study aims to eliminate the requirement for precise facial semantics and fine characteristics for the microphone object in face pictures by using a Generative Adversarial Network (GAN). It is split into two parts: the initial painting and the final polishing. Following the inpainter's examination of fine predictions, the refiner adds fine details underneath the microphone region by significantly filling it in, resulting in a more realistic image. The method was trained using artificial microphone pictures generated from CelebA face images, and it was subsequently assessed using actual microphone images collected throughout the study. A two-stage adversarial idea, EdgeConnect \cite{r12} consists of a network for image completion and an edge generator. The edge generator produces distorted edges for both regular and irregular missing regions, while the image completion network fills in the gaps produced by the edge generator's hallucinated edges. In short, this study first produces supervisory data, and then, in the second phase, it alters the picture of the subject. 

\vspace{-.8em}

\section{Methodology}
We divided the problem into three major modules: Landmark Module, Mask segmentation Module and Inpainting Module. Although we have another module named Gender Classification Module, which has been discussed in the latter part of the section.

\vspace{-.8em}
\subsection{LANDMARK PREDICTION MODULE}
The module for predicting landmarks $G_L(L_{gt})$ aims to predict 98 landmarks points from a masked face picture. The landmark prediction module is implemented in this work from \cite{r21}. To rebuild identified mask areas of an image, we are more concerned with the underlying topological structure and certain characteristics (position and expression) of the landmarks than with the exact location of each landmark. This landmark prediction model is built on Bulat et al. \cite{r24}'s stacked HG architecture. The  heatmap for each HG is supervised against the ground truth heatmap. For training, adaptive wing loss is employed, which may adjust its form to various kinds of ground truth heatmap pixels. This flexibility penalizes foreground loss more than background loss. To resolve the disparity between foreground and background pixels, developed the Weighted Loss Map, which gives a high weight to foreground and challenging background pixels, allowing the training process to concentrate on pixels critical for landmark localization.


\vspace{-.5em}

\subsection{MASK SEGMENTATION MODULE}
The mask segmentation module produces a binary segmentation map, ${M_m}$, with the mask object represented by 1 and the rest of the image's pixels represented by 0. The network topology for the mask segmentation module utilized in this research is based on the Mask R-CNN \cite{r29} structure. To be more specific, Mask R-CNN is a version of Faster R-CNN that is used to classify and identify various objects in an image. During object detection, different items in an image are given a class name and bounding box coordinates. Additionally, via semantic segmentation, Mask R-CNN generates an object mask for each item. Due to the presence of two phases, the model is divided into two halves. For object recognition, a Faster R-CNN architecture is utilized, whereas semantic segmentation is performed using a Fully Convolutional Network (FCN). Figure 2 illustrates the Mask R-CNN architecture and a more thorough process.

\begin{figure}[h]
\includegraphics[width=0.49\textwidth]{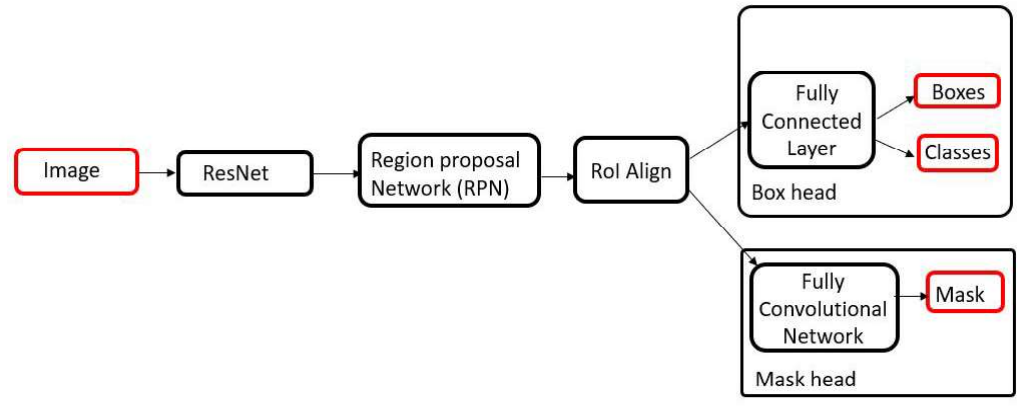}
\centering
\caption{Mask R-CNN Workflow.}
\end{figure}
\setlength{\textfloatsep}{6pt}
\begin{figure*}[h]
    \includegraphics[width=\textwidth, height=9cm]{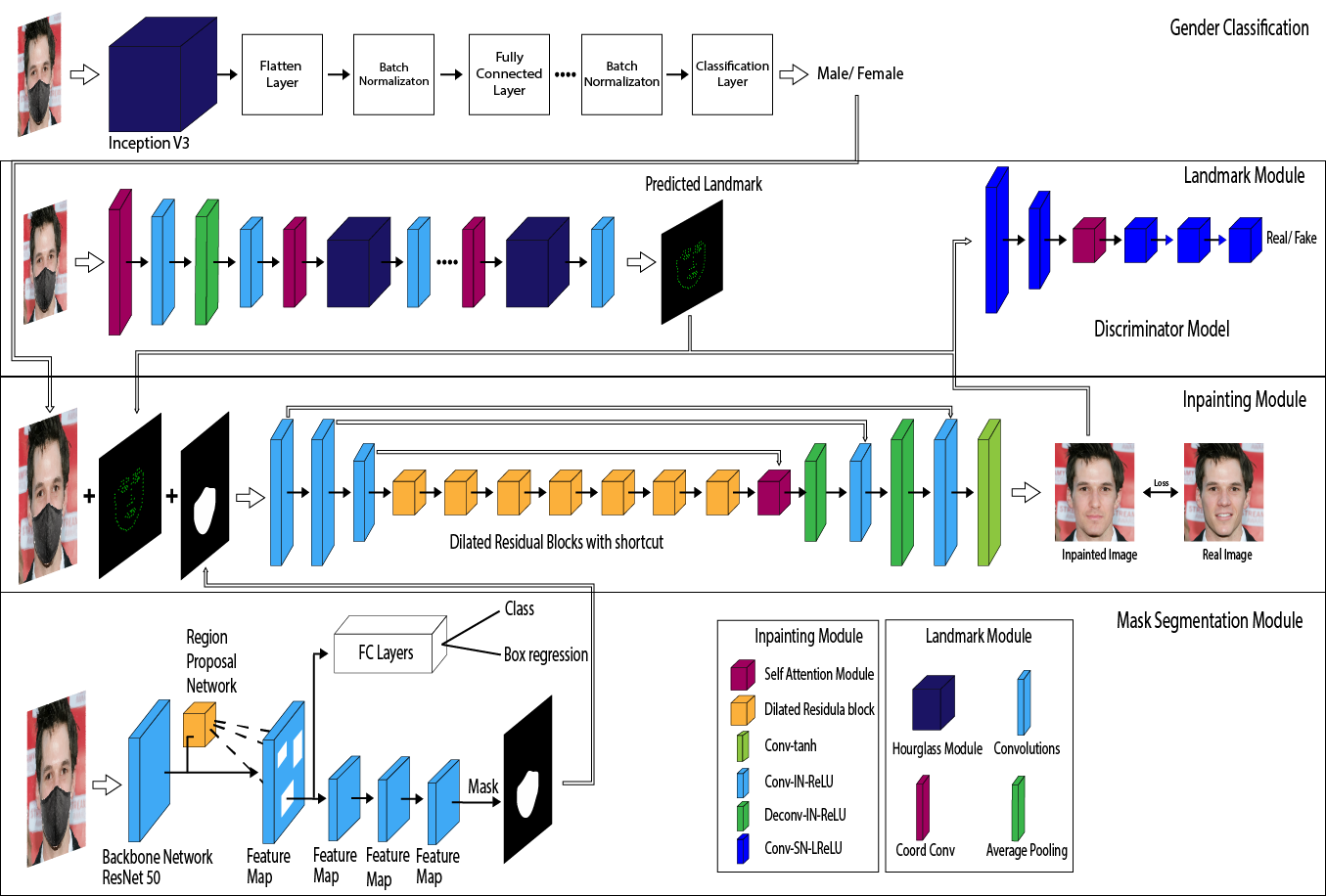}
    \centering
    \caption{The suggested model's network architecture. To begin, the landmark prediction module determines the location of a landmark in a face mask picture. The mask segmentation module then detects the mask area on the face. Following that, the inpainting module adds landmarks to the image before applying the given mask.}
    \end{figure*}
In this research, the ResNet50 network is used to extract features from images using Mask R-CNN, like the ConvNet network used in Faster R-CNN. This section's primary goal is to extract both low-level and high-level information from a picture. A Region Proposal Network is applied to the previously acquired feature map and predicts if an item is present in a given area or not. The 
Inception of the Union (IoU) with the ground truth box is computed for all predicted regions. When IoU is higher than or equal to 0.5, an area of interest is defined. Otherwise, it will languish in obscurity.


\vspace{-.8em}
\subsection{INPAINTING MODULE}
\setlength{\textfloatsep}{6pt}
The inpainting network $G_P$ seeks to complete faces by accepting as input the masked face images Igt and their (predicted or ground-truth) landmarks L ($L_p$ or $L_{gt}$), i.e. $I_p$:=$G_P$($I_{gt}$, L, $M_m$, $\theta_G$) with $\theta_G$ denoting the network parameters. The inpainting module consists of two models (Generator \& Discriminator).
\subsubsection{Generator}
The generator network is organized as a U-Net. More precisely, the network is composed of three blocks of progressively downsampled encoding, seven residual blocks with dilated convolutions, and a long-short term attention block. The feature maps are then progressively upsampled to the decoder's input resolution. The long-short attention layer \cite{r25} is used to link temporal feature maps, and stacked dilated blocks are utilized to expand the receptive field, allowing for the integration of more information.

Additionally, shortcuts are created between the relevant encoder and decoder layers. Additionally, the channel performs the 11 convolution operations before each decoding layer in order to change the weights of the shortcut and final layer features. As a result, the network may make more use of distant features on a geographical and temporal scale. The construction of the generator is shown in the inpainting module in figure 3.

\subsubsection{Discriminator}
In order to avoid being detected by the opponent, the generator constructs facial reconstructions by placing landmarks and relies on the fact that faces tend to vary even slightly from person to person. The discriminator then compares the facial reconstructions with the original image data to determine if they look similar. When the produced outcomes are indistinguishable from the actual ones, convergence occurs. This study uses the 70x70 Patch-GAN architecture \cite{r14} for our discriminator, D($I_{gt}$, L; $\theta_{D}$) 

with $\theta_{D}$ the parameters. We made the discriminator's training more stable by adding spectral normalization (SN) to its block \cite{r15}. Also, a layer dedicated to controlling the various functionalities is added. This work \cite{r5} used two discriminators: the global discriminator examines the picture in its entirety to judge its coherence, while the local discriminator just examines the recently finished area to make sure it matches the rest of the image. Conversely, our discriminator uses a single giver, who receives an image and its landmarks as input. There are three key explanations for this: The attention layer in the deep learning architecture works to promote attribute consistency. The fact that the produced results are reliant on the landmarks, which already offer global structure, besides we have trained the inpainting module separately for male and female which also offer gender attribute consistency, also contributes to a well-developed system. figure 3 displays our discriminator's setup.
\vspace{-1em}

\subsection{LOSS FUNCTION}
We train the inpainting model using a mix of style, per-pixel, perceptual, total variation, and adversarial loss.
\subsubsection{Style Loss}
The objective is to calculate a style matrix for both the produced and style images. This is because the root mean squares of the two style matrices will be used to determine the style loss. The style matrix is basically a Gram matrix, with the $j_{th}$ element calculated by multiplying the $i_{th}$ and $j_{th}$ feature mappings element-wise and summing over both width and height.Style loss is used to quantify the gap between two images.
\begin{align}
\nonumber
             \mathcal{L}_{style}:=\sum_{i} \frac  {1}{N_i * N_i} ||\frac{ G_p(I_p\circ M_m) - G_p(I_{gt} \circ M_m)}{N_i * H_i * W_i}||
\end{align}
where $\phi_p(x)$ equals  $\phi_p(x)^T \phi_p(x)$ denotes the Gram Matrix associated with $\phi_p(x)$.
\subsubsection{Per-pixel Loss}
Pixel-level differences between images are measured by using a loss function to evaluate each pixel. The mask size $N_m$ is used to adjust the penalty, which means that if there is any occulusion, $||.||$ evaluates $\ell$. The inpainted result should be near to the ground truth. When the level of corruption is excessive, structure and consistency are rational.
\vspace{-.8em}

\begin{align}
\nonumber
     \mathcal{L}_{pixel}:=\frac{1}{N_m}||I_p - I_{gt}||
\end{align}


\subsubsection{Perceptual Loss}
Before training, the perceptual loss is calculated using the VGG-19 $relu1_1, relu2_1, relu3_1, relu4_1,$ and $relu5_1$ on ImageNet \cite{r23}.The perceptual loss is a technique of measuring the clear distinction between convolution layers derived from a pre-trained network and that is defined as follow:
\begin{align}
\nonumber
     \mathcal{L}_{perc}:= \sum_{i} \frac{||\phi_i(I_p) - \phi_i(I_{gt})||}{N_i * H_i * W_i}
\end{align}
 
Np feature maps of size $H_i×W_i$, with $\phi_p(·)$ as the output from the p-th layer of the pre-trained network. To compute the perceptual loss and style loss mentioned below, we use relu1 1, relu2 1, relu3 1, relu4 1, and relu5 1 of the VGG-19 pre-trained on the ImageNet \cite{r23}.


\subsubsection{Total variation loss}
The complete variation loss is used to reduce what is known as the checkerboard artifact.
\begin{align}
\nonumber
    \mathcal{L}_{tv}:=\frac{1}{N_{I_{gt}}}||\nabla I_p||
\end{align}
 where ${N_{I_{gt}}}$ is the pixel count of I and is the first-order derivative including the terms $\nabla_h$ (horizontal) and $\nabla_v$ (vertical) (V). (V) Due to its stability throughout the training process and the improvement in visual quality, the adversarial loss employs the LSGAN described in \cite{r22}.

\subsubsection{Adversarial loss}
The consistency throughout the training period and the enhancement in visual quality. The adversarial loss employs the LSGAN described in \cite{r22}.
\begin{align}
\nonumber
    \mathcal{L}_{advG}:=\Xi[D(G_p(I_{gt},L,M_m),L_{gt})-1)^2]\\
\nonumber    
    \mathcal{L}_{advD}:=\Xi[D({I_p,L_{gt})^2})]+\Xi[D(I_{gt},L_{gt})-1)^2]
\end{align}

\subsubsection{Total Loss}
The overall loss in relation to the generator is equal to:
\vspace{-.8em}
\begin{multline*}
        \mathcal{L}_{inp}:= \mathcal{L}_{pixel}+	\lambda_{perc}\mathcal{L}_{perc}+\lambda_{style}\mathcal{L}_{style}+\\\lambda_{tv}\mathcal{L}_{tv}+\lambda_{adv}\mathcal{L}_{advG}
\end{multline*}
In our trials, we utilize  $\mathcal{L}_{perc}$ = 0.1,  $\mathcal{L}_{style}$ = 250, $\mathcal{L}_{tv}$ = 0.1, and $\lambda_{adv}$ = 0.01. Throughout the training process, $\mathcal{L}_{inp}$ for the generator $G_p$ and $\mathcal{L}_{advD}$ for the discriminator D are minimized alternately until convergence.

\vspace{-.8em}

\subsection{GENDER CLASSIFICATION WITH FACE MASK MODULE}
We utilized inceptionV3 to develop our gender categorization model in this study. InceptionV3 was trained using the ImageNet dataset. We used this model to extract features from images. We have done fine-tuning in this model and added custom fully connected layers with custom node numbers.

\begin{figure}[h]
\includegraphics[width=0.49\textwidth]{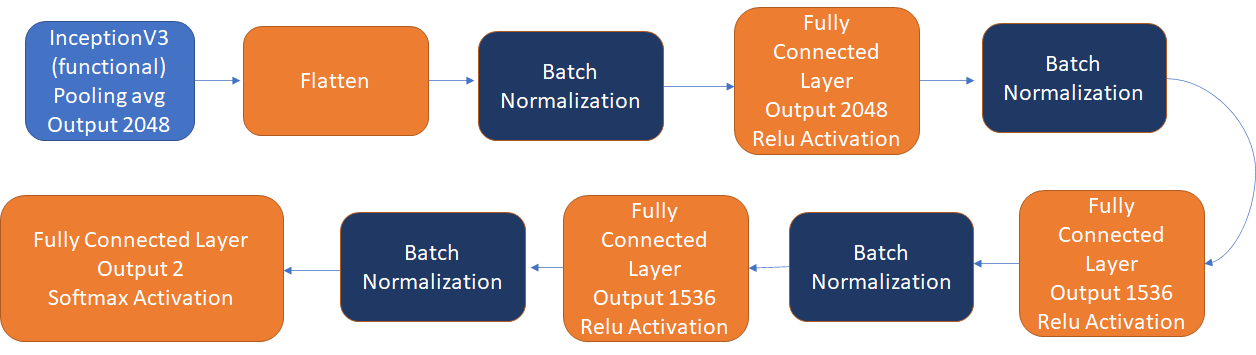}
\centering
\caption{Gender Classification Model.}
\end{figure}

As shown in Figure 4, we built the custom fully connected layers. In each densely connected hidden layer, we utilized the Relu activation function, whereas, in the output layer, we used the Sigmoid activation function. After each layer, we used Batch Normalization.

\section{Experiments}

\subsection{SYNTHETIC DATASET}
\begin{figure}[t]
\includegraphics[width=0.53\textwidth]{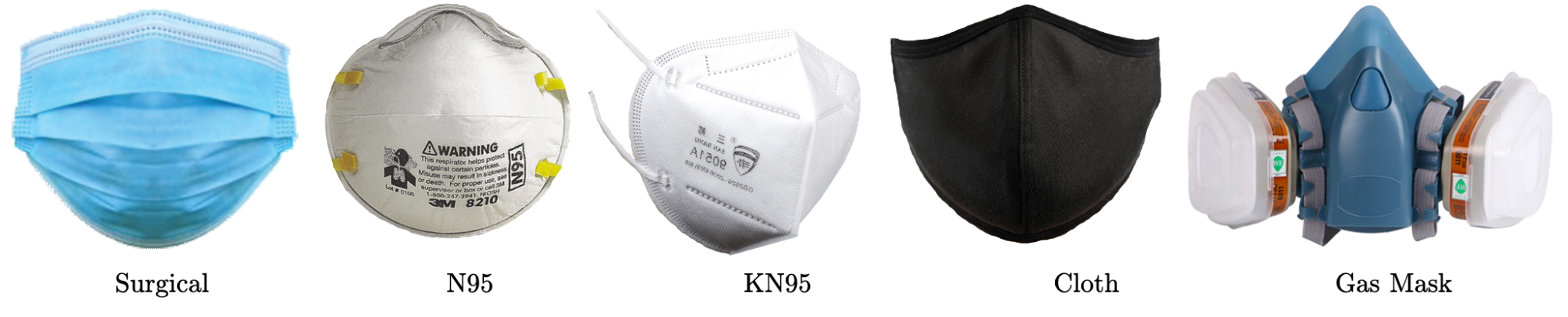}
\centering
\caption{An illustration of the many types of masks used in the training dataset.}
\vspace{+.5em}
\end{figure}
No publicly accessible dataset containing face picture pairings with and without mask objects exists to allow us to train our model in an unsupervised way using the information from this dataset. We created a synthetic dataset using the Flickr-Faces-HQ Dataset (FFHQ), which is freely accessible on the internet \cite{r20}. The dataset includes 52,000 high-quality PNG pictures with a size of 512*512 pixels and has a great deal of diversity in terms of age, ethnicity, and the backdrop of the photographs. It also offers extensive coverage of accessories such as eyeglasses, sunglasses, caps, and other similar items of clothing. We began by dividing the dataset into two subsets (male and female) using the gender categorization model to separate the data. In our synthetic dataset, we utilized a variety of different types of masks with varying sizes, forms, colors, and structural characteristics.
\begin{figure}[h]
\includegraphics[scale=0.8]{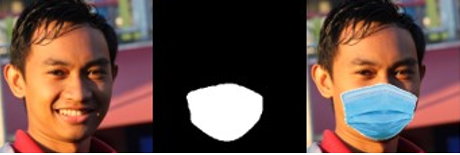}
\centering
\caption{Sample images of the training dataset}
\end{figure}
 \vspace{-1em}
Figure 5 depicts some of the face masks found in our dataset, as well as some of their characteristics. For the purpose of creating synthetic samples, we used Face MasedNet \cite{r27} to apply masks to a face while simultaneously generating the appropriate binary segmentation mask for the mask applied. A picture from our synthetic dataset is shown in Figure 6. 

\vspace{-.8em}

\subsection{MASK SEGMENTATION DATASET}
For getting better results of mask segmentation for real-world images, we have collected face images with masks using Google scraping. The dataset consists of 500 images with different kinds of masks of different sizes, shapes, colors and structures. we preprocess the image, by resizing the images into 256x256. We have used the Labelme tool \cite{r26} to label the mask region by hand. For each image, there is a custom JSON file where we have stored the annotation values.
\vspace{-.8em}

\subsection{GENDER CLASSIFICATION DATASET}
The gender classification dataset used in this paper is FFHQ, while the remaining pictures are taken from Google scraping. There are 89,443 pictures in total for training, which are divided into two groups (male and female). Additionally, there are 20,714 pictures for validation. First, the data prepossessing is performed where we have classified those images in FFHQ into two classes manually, resizing the images into 256x256. We utilized Face MaskedNet\cite{r27} to apply masks to a face.

\vspace{-.8em}

\subsection{TRAINING DETAILS}
\subsubsection{Landmark Module}
We discovered the AdaptiveWingloss landmark module \cite{r21} which provides good results with a significant obstruction. Following that, we conducted tests using our dataset and conducted research using the findings. We obtained the anticipated outcomes. As a consequence, we utilized their pre-trained model to determine the location of a landmark from an image.

\subsubsection{Mask Segmentation Module}
In this lesson, we trained our mask dataset using the Mask R-CNN model. Due to the limited size of our mask dataset. As a consequence, we were forced to use the method of transfer learning. As a result, we utilized the COCO dataset's pre-trained weights, which were learned using the Mask R-CNN model. The Mask R-CNN model was trained using 256x256 images. The model was trained using a batch size of two and a learning rate of \num{2.5e-4} for ten thousand iterations. We ended up with a loss of 0.0878.

\subsubsection{Inpainting Module}
We utilized the FFHQ\cite{r20} \& CelebA\cite{r30} dataset to train the inpainting module's generator and discriminator. We began by resizing the pictures to 256x256. We conducted many tests with the inpainting model and discovered a gap in correctly preserving the gender property. As a result, we split the FFHQ \& CelebA dataset into male and female subsets. Then, we trained distinct datasets for males and females and optimized them using the Adam optimizer \cite{r16} with $\beta_1 = 0$ and $\beta_2$ = 0.9 and a learning rate of \num{10e-4}. The discriminator has a learning rate of \num{10e-5}. We trained the model for 500000 iterations with a batch size of 4 for both subsets.

\begin{figure}[h]
\includegraphics[width=0.49\textwidth]{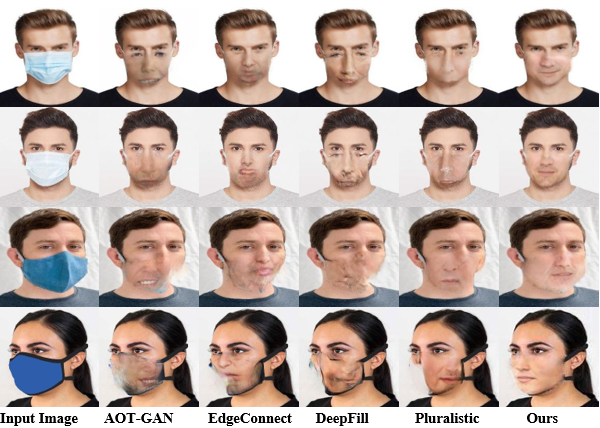}
\centering
\caption{Real world classified images results for AOT-GAN, EdgeConnect, Deepfill, Pluralistic, and Ours. The shown result of Our model which is trained using celebA dataset.}
\end{figure}

\subsubsection{Gender Classification with Face Mask Module}
We used the pre-trained weight of the InceptionV3 which was trained using ImageNet and constrained the weights untrainable. We trained the custom fully connected layer. There are 9,720,321 trainable parameters in all. The input tensor of InceptionV3\cite{r31} is set to (256, 256, 3). The Adam optimizer is used in the experiment, with binary cross-entropy as the loss function, 256 batches, and a learning rate of 1e-4. Furthermore, the number epoch is limited to 150. We have got an accuracy of 88.12 percent over the test data. 

\begin{figure}[h]
\includegraphics[width=0.49\textwidth]{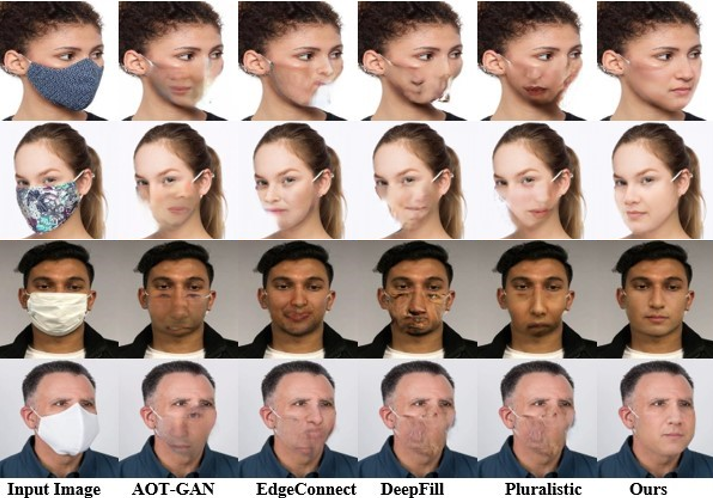}
\centering
\caption{Real world classified images results for AOT-GAN, EdgeConnect, Deepfill, Pluralistic, and Ours. The shown result of Our model which is trained using FFHQ dataset.}
\end{figure}
\setlength{\textfloatsep}{6pt}

\vspace{-1em}

\subsection{QUALITATIVE RESULT}
In this part, we evaluate the results produced by our model on real-world test images and compare them to the results produced by other state-of-the-art image editing methods such as DeepFill \cite{r17}, Pluralistic \cite{r13}, EdgeConnect \cite{r12}, and AOT-GAN \cite{r18}.

Figure 1 depicts a sample of actual test pictures produced by our model, as shown in the previous section. Our test samples exhibit a wide range of features, including the size, shape, color, and structure of the masks. A mask obscures almost half of the most noticeable facial features in each of the test pictures. As shown in Figure 10, the mask segmentation model is effective in removing the mask object and generating outputs that are natural in appearance while maintaining structural integrity. Figures 7 and 8 illustrate our model in comparison to other state-of-the-art techniques. The results show that our approach is successful at eliminating the mask object and completing the face, which seems not only structurally cohesive but also physically realistic. Pluralistic \cite{r13}, EdgeConnect \cite{r12}, DeepFill \cite{r17}, and AOT-GAN \cite{r18} are all unable to accomplish the job in the proper manner. This study has previously been discussed in detail in the section on related studies. Pluralistic generates convincing new output, but when applied to face images with missing areas and significant structural and appearance differences, it fails to create plausible results. The contextual attention layer is at the core of the DeepFill technique since it learns to build missing patches by copying feature information from known background patches. 
\vspace{-1em}
\setlength{\textfloatsep}{6pt}
\begin{figure}[h]
\includegraphics[width=0.49\textwidth]{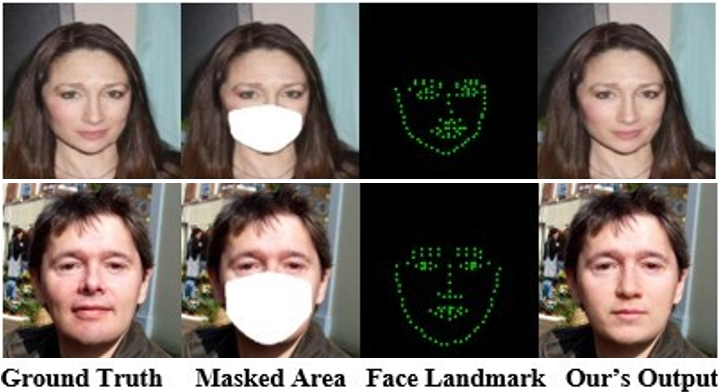}
\centering
\caption{Performance for known Ground Truth image for FFHQ (1st row) \& CelebA(2nd row) Dataset trained model.}
\end{figure}
\vspace{-.5em}
It works well for images when the probability of finding similar patterns in adjacent patches is high but fails miserably when dealing with large missing regions. On the other hand, the final output of Edgeconnect is reliant on the edge map generated in the first phase. When confronted with this problem, where the missing region is large, the edge generator is unable to generate appropriate edges, and as a consequence, the EdgeConnect network's final output suffers as well. When using AOT-GAN, just the missing area is generated; the remainder of the picture is left in its original state. As a result, it creates the missing semantics correctly, but it can’t maintain the actual shape and facial features of a face, and it generates multiple facial features (more than two eyes, lips) for a single face, as seen in the image above. As can be seen, we have obtained more accurate and realistic results for various facial angles. Additionally, we have the real gender output, which contains the precise facial characteristics of both males and females.
{\href{https://drive.google.com/drive/folders/1RAcxpuBsmj8muouTK3NxlhAdcYvYTPT5?usp=sharing}{\textbf{The datasets are available on: shorturl.at/osuKU}}}

\vspace{-1em}


\subsection{QUANTITATIVE SECTION}
In the PSNR, the higher the value of the metric, the more efficient it will be, and in the case of SSIM, the closer to 1, the better the quality image will be. \textbf{As we have trained our model separately for male and female, we have got the matric results for both. That's why, we are unable to compare our quantitative results to those of other similar research studies where they did not follow our's methodology.}
\\
\\
\begin{tabularx}{0.49\textwidth}{ 
  | >{\centering\arraybackslash}X 
  | >{\centering\arraybackslash}X 
  | >{\centering\arraybackslash}X 
  | >{\centering\arraybackslash}X | }
 \hline
 Dataset & Metric & Male & Female \\
 \hline
 \multirow{2}{4em}{FFHQ} & PSNR &  28.82 & 30.00  \\
 & SSIM &	0.93 &	0.95  \\
 \hline
 \multirow{2}{4em}{CelebA} & PSNR &  31.40 & 33.32  \\
 & SSIM &	0.97 &	0.98  \\
 \hline
\end{tabularx}

\vspace{-.5em}

\section{Conclusion}
As a result of this research, we have developed a framework for properly inpainting a masked face. Reconstruction of the face after mask removal is a difficult task because we must deal with a variety of issues, including mask edges on faces after mask removal, different mask shapes, and variable angles of face images. Thus, we used GAN-based image inpainting in conjunction with the image-to-image translation technique to get plausible results for comprehensive and natural image reconstruction. To increase realism, we trained distinct models for males and females, which addressed issues with picture inpainting where earlier attempts were unable to identify key facial features between males and females.
Our guiding concept is that the landmarks should be sufficiently tidy, adequate, and sturdy to act as a guide for the face inpainting module when it comes to supplying structural information. We developed a technique for collecting remote spatial context and linking temporal feature maps together to guarantee attribute consistency. Because of the usage of landmarks, our network becomes more adaptable, and regardless of the shapes we get, we can reconstruct the image with perfect form. The validity of our claims, as well as the efficacy of our design, have been established through extensive experiments that have demonstrated both qualitative and quantitative improvements over the current state of the art alternative designs. When compared to existing state-of-the-art image editing techniques, the findings of qualitative comparisons indicate that our model has produced high-perceptual-quality outcomes in the case of significant missing gaps in face pictures when compared to the results of previous studies.

\vspace{-.8em}


%


\ifCLASSOPTIONcaptionsoff
  \newpage
\fi

\bibliographystyle{unsrt}
 \bibliography{references}


\end{document}